\documentclass[a4paper,fleqn]{cas-dc}

\usepackage[numbers]{natbib}
\usepackage{soul}

\makeatletter
\AtBeginDocument{\let\hl\@firstofone}
\makeatother

\def\tsc#1{\csdef{#1}{\textsc{\lowercase{#1}}\xspace}}
\tsc{WGM}
\tsc{QE}
\tsc{EP}
\tsc{PMS}
\tsc{BEC}
\tsc{DE}
\begin{document}
\let\WriteBookmarks\relax
\def\floatpagepagefraction{1}
\def\textpagefraction{.001}
\shorttitle{Learning Pose-invariant 3D Object}
\shortauthors{Bo Peng et~al.}

\title [mode = title]{Learning Pose-invariant 3D Object Reconstruction from Single-view Images}                      
% \tnotemark[1,2]

% \tnotetext[1]{This document is the results of the research
%   project funded by the National Science Foundation.}

% \tnotetext[2]{The second title footnote which is a longer text matter
%   to fill through the whole text width and overflow into
%   another line in the footnotes area of the first page.}

\author[1]{Bo Peng} %[type=editor,
                        % auid=000,bioid=1,
                        % prefix=Sir,
                        % role=Researcher,
                        % orcid=0000-0001-7511-2910]
% \cormark[1]
% \fnmark[1]
\ead{bo.peng@nlpr.ia.ac.cn}
% \ead[url]{www.cvr.cc, cvr@sayahna.org}

% \credit{Conceptualization of this study, Methodology, Software}

\address[1]{Center for Research on Intelligent Perception and Computing (CRIPAC), \\
National Laboratory of Pattern Recognition (NLPR), \\
Institute of Automation Chinese Academy of Sciences (CASIA)}

% \address[2]{Shenzhen University}

\author[1]{Wei Wang} %[style=chinese]
\ead{wwang@nlpr.ia.ac.cn}

\author[1]{Jing Dong} %[%
%   role=Co-ordinator,
%   suffix=Jr,
%   ]
% \fnmark[2]
\ead{jdong@nlpr.ia.ac.cn}
% \ead[URL]{www.sayahna.org}

% \credit{Data curation, Writing - Original draft preparation}

\author%
[1]
{Tieniu Tan}
% \cormark[2]
% \fnmark[1,3]
\ead{tnt@nlpr.ia.ac.cn}
% \ead[URL]{www.stmdocs.in}

\cortext[cor1]{Corresponding author}
\cortext[cor2]{Principal corresponding author}
\fntext[fn1]{This is the first author footnote. but is common to third
  author as well.}
\fntext[fn2]{Another author footnote, this is a very long footnote and
  it should be a really long footnote. But this footnote is not yet
  sufficiently long enough to make two lines of footnote text.}

\nonumnote{This note has no numbers. In this work we demonstrate $a_b$
  the formation Y\_1 of a new type of polariton on the interface
  between a cuprous oxide slab and a polystyrene micro-sphere placed
  on the slab.
  }

\begin{abstract}
Learning to reconstruct 3D shapes using 2D images is an active research topic, with benefits of not requiring expensive 3D data. However, most work in this direction requires multi-view images for each object instance as training supervision, which oftentimes does not apply in practice. In this paper, we relax the common multi-view assumption and explore a more challenging yet more realistic setup of learning 3D shape from only single-view images. The major difficulty lies in insufficient constraints that can be provided by single view images, which leads to the problem of pose entanglement in learned shape space. As a result, reconstructed shapes vary along input pose and have poor accuracy. We address this problem by taking a novel domain adaptation perspective, and propose an effective adversarial domain confusion method to learn pose-disentangled compact shape space. Experiments on single-view reconstruction show effectiveness in solving pose entanglement, \hl{and the proposed method achieves on-par reconstruction accuracy with state-of-the-art with higher efficiency.}
\end{abstract}

\begin{graphicalabstract}
\includegraphics{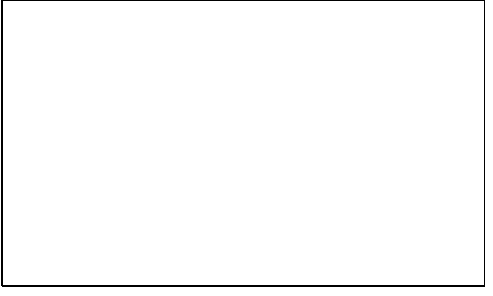}
\end{graphicalabstract}

\begin{highlights}
\item Propose a learning based method to reconstruct 3D object shapes using only single view images.
\item Tackle the problem of pose entanglement by devising an adversarial domain confusion method.
\item Justify the superiority in both reconstruction accuracy and efficiency on two benckmark datasets.
\end{highlights}

\begin{keywords}
Learning 3D shape \sep Single view supervision \sep Domain confusion \sep Adversarial learning
\end{keywords}

\maketitle

%% main text
\section{Introduction}

Inferring 3D shape of an object from image is a long-standing fundamental problem of computer vision. Although accurate geometry information can be reconstructed from multiple views of a scene using stereo matching or structure from motion methods, machines still can not reliably reconstruct high quality shapes from single view images like humans do. This is primarily hindered by ill-posedness of the problem, thus prior knowledge of 3D shapes is required. With the success of deep neural networks, more and more work tries to learn 3D shape priors \cite{wu2016learning} from 3D data or directly learns the mapping from 2D image to 3D shape \cite{wang2018pixel2mesh,fan2017point}. However, these methods require large datasets of 3D models, which is costly and sometimes even impossible. Compared to 3D shape data, images are more common and easy to capture. As a result, there is a ongoing interest for learning 3D shape models from only 2D images.

\begin{figure}[t]
\centering
\centerline{\includegraphics[width=8cm]{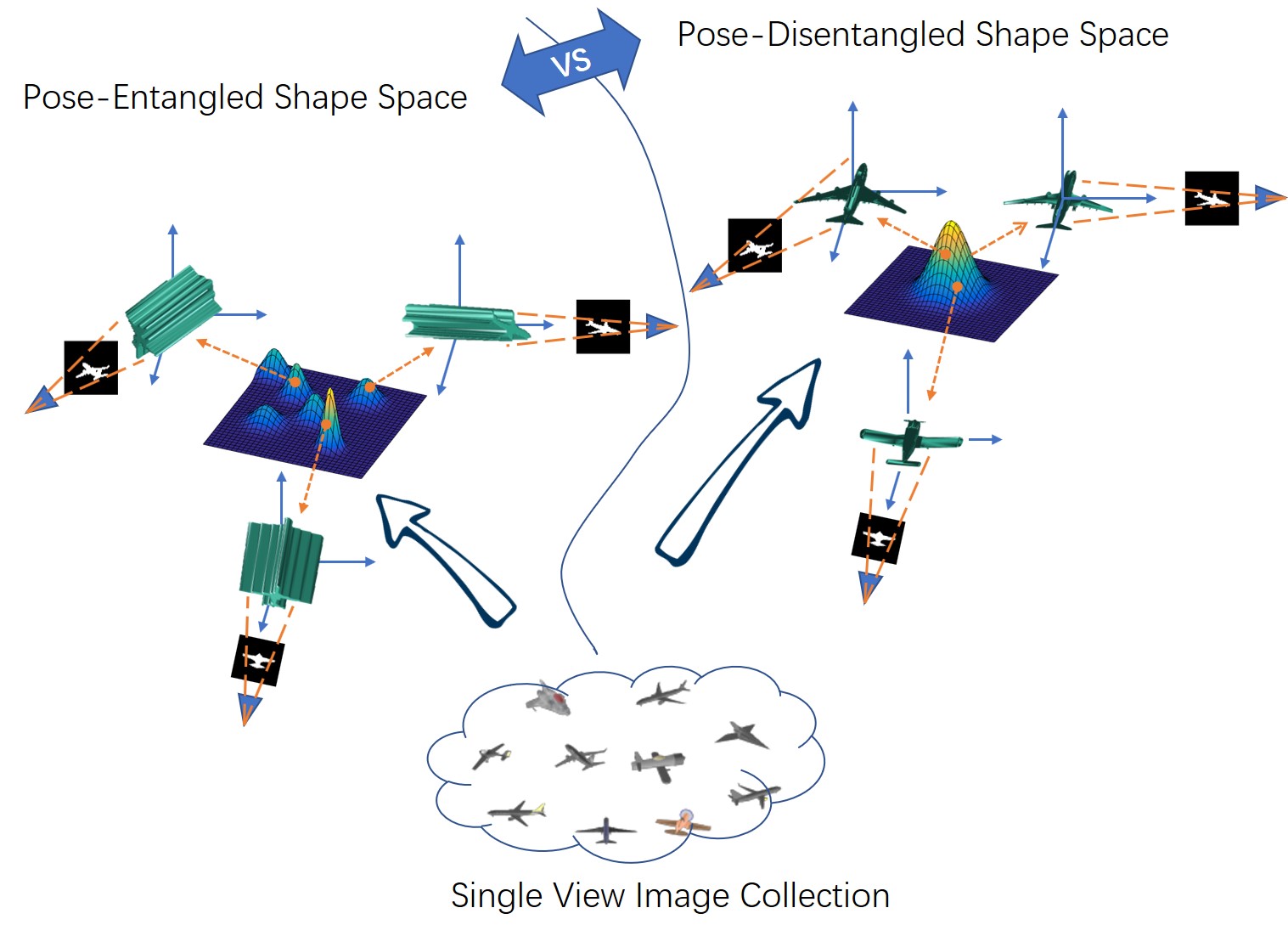}}
\caption{In this work, we focus on learning 3D shape model from single view images. A challenging pitfall is that it tends to learn a pose-entangled shape space that has multiple modes of unlikely shapes which only project correctly under certain input camera pose (e.g. the relief style shapes on the left). According to Occam's razor theory, the pose-disentangled shape space on the right is preferred, since it is the most simple model that explains all image observations. We propose to learn this model by adding explicit constraints to pull multiple pose modes together in the shape embeding space.}
\label{figIllustration}
\end{figure}

Almost all existing deep-learning based 3D reconstruction methods that use 2D images as supervision require multi-view images of each object instance, e.g.  \cite{yan2016perspective,liu2019soft,tulsiani2017multi,yang2018learning,gwak2017weakly}. This is because 2D supervision is much weaker compared to direct 3D supervision, and there exists infinitely many 3D shapes that can explain a given single-view image observation. Hence, researchers resort to multi-view images to constrain solution space. Yan et al. \cite{yan2016perspective} are among the first to explore multi-view supervised shape learning, where they used $24$ different views of each object instance with known pose annotation. In later works, different methods are proposed using either $5$ views \cite{tulsiani2017multi} or as least as $2$ views \cite{kato2018neural,tulsiani2018multi,insafutdinov2018unsupervised} with either known or estimated pose information. We argue that in practice, multi-view image datasets are still costly, since it demands extra labor of annotation and compilation. On the contrary, unstructured single-view images are more common and readily available, and they serve as a more convenient data resource for learning 3D shape models.

In this work, we target the more practical yet challenging task of learning 3D shapes from only single-view images. We show that without supervision from multiple different viewpoints, this problem becomes very hard, even with the knowledge of pose information. The major challenge is found to be what we call the pose entanglement, which is the problem of shape embeddings being entangled with poses. As a result, reconstructed shapes vary greatly with different input viewpoints. See the illustration in Fig. \ref{figIllustration} for more intuitive understanding. Note that object's viewpoint and camera pose are two equivalent concepts here.

We propose a deep auto-encoder based network that learns 3D mesh models from single-view images in a self-supervised reconstruction-projection-and-check manner. Our key contribution for addressing the problem of pose entanglement is to cast it in a domain adaptation perspective, where shape embeddings from different viewpoints are treated as different domains that are pulled together to the same distribution. The motivation is that the underlying 3D shapes that give raise to diverse observations in different viewpoints should form a single compact space.
Accordingly, we propose a novel adversarial domain confusion loss and train a pose discriminator in shape embedding space. 
\hl{Different from the previous work \mbox{\cite{kato2019learning}} that tackles the same problem using adversarial training in the re-projected image domain, our direct domain confusion training in the shape embedding domain is a more elegant simplification. The proposed method is competitive with state-of-the-art and more efficient.}
Comprehensive experiments on single image reconstruction show very promising results and demonstrate that the proposed model indeed learns a pose-disentangled and compact shape space.
\textit{\textbf{We also make our source code publicly available for reproducible research.}}\footnote{ \url{https://github.com/bomb2peng/learn3D}}

\section{Related Work}
Learning 3D shape model is an actively studied area, where there exists multiple choices to represent a 3D model, such as voxels, point clouds, meshes and combination of geometry primitives. Each 3D representation has its pros and cons in aspects of intuitiveness, complexity, and accuracy. In this work, we choose 3D mesh, which is a natural and complete representation for object surfaces.
In the following, we overview related work from the aspects of 3D or 2D supervision, multi-view or single-view supervision and different levels of annotation.
\\

% \textbf{3D representation.} Many different representations for 3D model are explored in previous work, including voxels, point clouds, meshes, geometry primitives and 2D representations. They have different pros and cons in aspects of intuitiveness, implementation complexity, efficiency and accuracy. 3D voxel is probably the most wildly used representation for 3D model learning \cite{wu2016learning},  because of its easy integration with convolutional neural networks. However, 3D voxels have the problem of low efficiency, leading to operating on relatively low resolution grids. Point clouds \cite{fan2017point} only represent 3D model surfaces and is hence more efficient, but it is not a complete surface representation which lacks normal information and can not calculate surface shading. Mesh consists of 3D vertices and edges \cite{kato2018neural}, making it a natural and complete representation for object surfaces. Although widely used in computer graphics field, meshes are more complex to model. Geometry primitives like compositions of cuboids are easy to implement and learn, which has very few parameters, but this representation is very coarse \cite{abstractionTulsiani17}. Various 2D representations represent 3D surfaces as 2D images, e.g. UV representation \cite{tran2018learning} and Spherical Map \cite{genre}, which makes the usage of normal 2D CNNs possible. In this work, we use the natural 3D mesh representation and add smoothness regularization to prevent unsmoothness.
% \\

\hl{\textbf{Learning with 3D supervision.} }
% Early 3D morphable models (3DMMs) \cite{blanz1999morphable,paysan20093d} are statistical shape models built on 3D mesh data using PCA. 3DMMs are very useful for generating plausible shapes, notably for faces. By fitting to image observations, it can reconstruct 3D shape from a single image.
% However 3DMMs are hard to build requiring accurate 3D registration and data pre-processing.
Recently, learning based methods develop fast.  Generative adversarial network (GAN) \cite{goodfellow2014generative} is applied on 3D voxel data in 3DGAN \cite{wu2016learning} to learn 3D shape distributions. 3DGAN is unsupervised learned and can be applied to single-view reconstruction by additionally training an inference network post-hoc. \hl{In a similar spirit, the work in \mbox{\cite{learning_IJCV20}} proposes a weakly supervised 3D shape completion method, which uses synthetic 3D data to train a VAE as shape prior and amortize maximum likelihood fitting using deep neural networks.}

Some other works learn single-view 3D reconstruction \cite{wang2018pixel2mesh,fan2017point,wu2017marrnet,papier_CVPR18,occupancy_CVPR19} or multi-view 3D reconstruction using direct 3D model supervision \cite{3d_ECCV16,robust_IJCV20}. For single-view reconstruction, apart from commonly used 3D representations like mesh \cite{wang2018pixel2mesh}, point cloud \cite{fan2017point} and voxel \cite{wu2017marrnet}, some works propose novel representations \cite{papier_CVPR18,occupancy_CVPR19}. \hl{AtlasNet \mbox{\cite{papier_CVPR18}} proposes a representation of a collection of parametric surface elements called Atlas. This flexible representation enables both 3D surface reconstruction from point clouds and from single-view image with better precision and generalization. The work in \mbox{\cite{occupancy_CVPR19}} on the other hand proposes a new representation named occupancy network that represents 3D surface as continuous decision boundary of deep neural network classifier.}

\hl{For multi-view reconstruction, in \mbox{\cite{3d_ECCV16}}, authors propose a novel 3D recurrent reconstruction neural network (3D-R2N2) that reconstructs 3D voxel model from one or more input images. The work \mbox{\cite{robust_IJCV20}} proposes an attention module and a dedicated training algorithm to robustly aggregate deep features extracted from arbitrary number of input view images for multi-view reconstruction.}
The 3D supervised methods achieve very good results on 3D reconstruction, but it requires expensive image-model pairs that are hard to acquire for real-world images.
\\

\hl{\textbf{Learning with multi-view supervision.}} Since collecting 2D images is more affordable compared to collecting 3D models, much research focuses on learning 3D shape model using multi-view images as supervision.
% \cite{yan2016perspective,kato2018neural,liu2019soft,tulsiani2017multi,tulsiani2018multi,insafutdinov2018unsupervised,yang2018learning,gwak2017weakly}.
Image observations of the same object instance from multiple viewpoints add constraints for this problem and make it easier to solve. Differentiable projection modules are proposed in these works to bridge 3D model and 2D projection, and 3D models are learned through minimizing image re-projection loss under multiple views. Silhouette is the mostly used 2D supervision because of its simplicity and robustness to lighting and texture. In this line of work, Yan et al. \cite{yan2016perspective} use 24-view silhouettes as supervision to train an encoder-decoder model of voxel reconstruction. Similarly, Tulsiani et al. \cite{tulsiani2017multi} reconstruct 3D voxels with more general 2D supervisions, such as depth, color images and semantic labels. Gwak et al. \cite{gwak2017weakly} additionally use adversarial constraints learned from 3D data to get better reconstruction results. Mesh models can also be learned  \cite{kato2018neural,liu2019soft} with the help of mesh renderers. 

\hl{RenderNet \mbox{\cite{rendernet_NIPS18}} proposes a differentiable rendering CNN with novel projection unit, where complex rendering effects are learnt by network. It can be used for inverse rendering by iterative optimization of shape and lighting etc. under multi-view. DeepMVS \mbox{\cite{deepmvs_CVPR18}} designs a CNN for multi-view stereo reconstruction, benefiting from synthetic data training and aggregating information over unordered image set. The work in \mbox{\cite{scene_NIPS19}} propose Scene Representation Networks (SRN) as a continuous 3D structure-aware scene representation that encodes both geometry and appearance. With the priors learnt by self-supervision, SRN can better reconstruct 3D model from multi-view posed images. NeRF \mbox{\cite{nerf_arxiv20}} presents a method for synthesizing novel views by optimizing an underlying continuous volumetric scene function using a sparse set of input views.}

The above works require known poses (or viewpoints) to learn the model. To relax pose annotation, in \cite{tulsiani2018multi,insafutdinov2018unsupervised} both shape and pose are inferred simultaneously to ensure cross-view projection consistency. In \cite{yang2018learning}, images with and without pose annotation are combined to learn shape model using both re-projection loss and adversarial loss. All the above methods require multi-view supervisions and hence are limited in applicable range.
\\

\hl{\textbf{Learning with single-view supervision.}} Some traditional computer vision methods exist for learning shape models from single-view images. Cashman et al. \cite{cashman2012shape} propose to build 3D morphable models (3DMMs) \cite{blanz1999morphable,paysan20093d} from single-view image collections with both keypoint and silhouette annotations. In \cite{kar2015category}, a similar 3DMM learning strategy is designed by firstly using non-rigid structure-from-motion for estimating poses across the image set. Different from these linear morphable models, recently more powerful shape models in the form of neural networks are proposed and learned from single-view image collections of birds \cite{kanazawa2018learning} and faces \cite{tran2018learning,tran2018nonlinear}. However, these methods need multiple annotations like silhouettes and keypoints, and they cannot cold start and need careful initialization operations.

\hl{Henderson and Ferrari propose VAE based generative modeling of shape, pose and shading \mbox{\cite{henderson2019learning}}, and they learn to infer these factors by training on single-view image dataset. Their disentanglement of all three factors from a single-view image is a great achievement, but their experiments are all conducted in controlled synthetic setting where objects are assumed texture-less and lighting conditions are pre-fixed, which is still rather restricted for real-world application. Later, the same author proposes a generative model of textured 3D meshes \mbox{\cite{leveraging_CVPR20}}, also in the form of VAE, learned from weakly annotated image datasets. This work mainly focuses on realistic texture modeling, no evaluation on shape reconstruction accuracy is conducted, and it also assumes weak supervision in the form of camera calibration and masks. The work in \mbox{\cite{unsupervised_CVPR20}} proposes unsupervised learning of probably symmetric 3D models from in the wild image sets, exploring the effective symmetry constraints. This method mainly targets on near frontal surfaces, and will have trouble reconstructing full 3D objects because of unobserved parts.}

Another work \cite{gadelha20173d} uses GAN on 2D projections to learn generative 3D shape model whose silhouette projections are indistinguishable from those of real shapes. However, this model is not directly trained to infer 3D shape from image, and our adversarial training is different from this work, since ours focus on shape embedding space. To our best knowledge, the most closely related work is the view-prior-learning model (VPL) in \cite{kato2019learning} which tackles the same problem as ours. While both work uses adversarial training to obtain pose-invariant shape reconstruction, the VPL model \cite{kato2019learning} conducts adversarial training on re-projected image domain while ours focuses on shape embedding domain. This makes our method more direct and intuitive compared to \cite{kato2019learning}, \hl{and we also provide experimental comparisons proving our competitive accuracy and higher efficiency.}

\section{Proposed Model}
\label{secModel}
\begin{figure}[htb]
\centering
\centerline{\includegraphics[width=9cm]{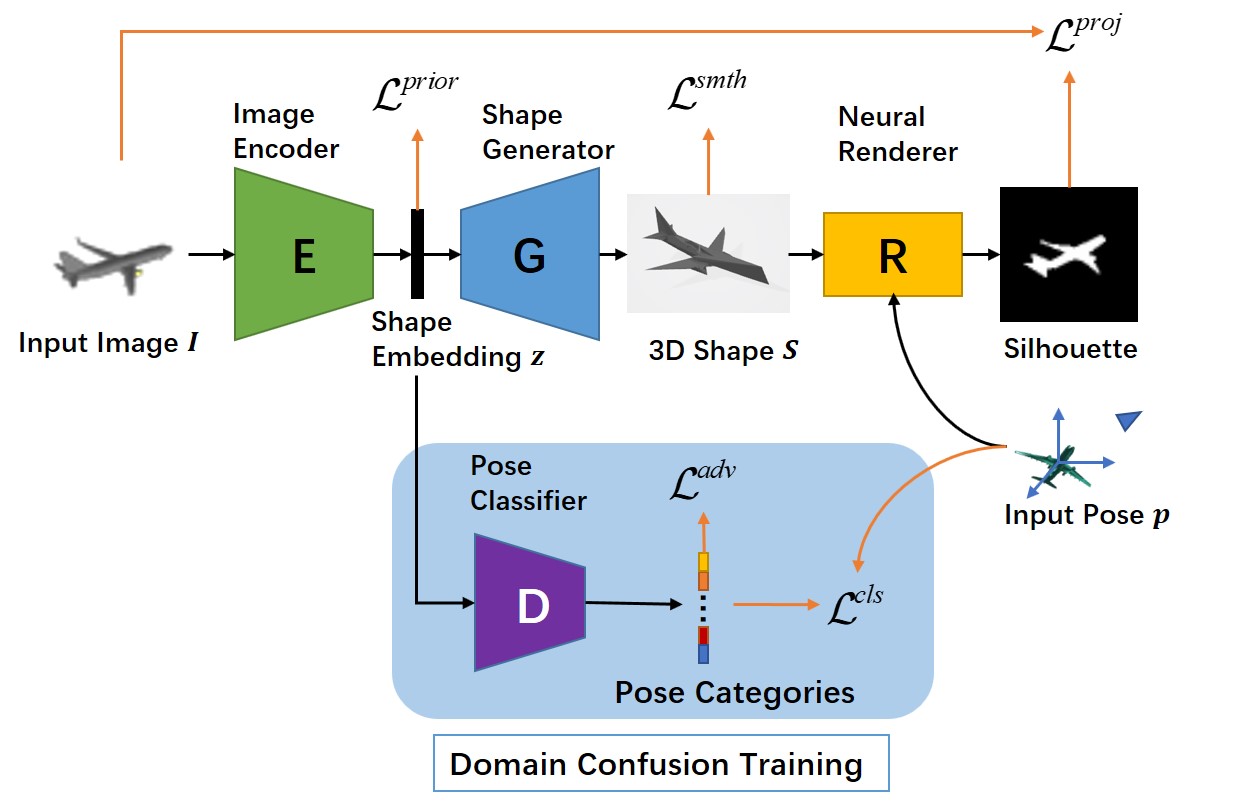}}
\caption{The overall structure of proposed model. See text for detailed descriptions.}
\label{figFramework}
\end{figure}

The overall proposed model is shown in Fig. \ref{figFramework}. It is a generative model of 3D mesh, which is an efficient and commonly-used 3D shape representation. It is based on encoder-decoder structure, where input image is encoded into shape codes and decoded to 3D mesh model, which is subsequently re-projected to 2D silhouette by a differentiable mesh renderer under given pose parameters and then checked against the true silhouette of input image. To tackle the problem of pose entanglement, we also propose a domain confusion module and add prior regularization on shape codes, which drives the model to learn pose-disentangled and compact 3D shape space. More details are presented in the following.

\textbf{Problem Formulation.} Given a set of images $\{I_i^C\}$ of a specific category of rigid object $C$, assuming object poses are known as $\{ p_i \}$, we want to learn a neural network model $G(z)$ that can generate plausible 3D shapes of $C$, where $z$ represents shape code. In this work, we consider finite categories of discrete poses for simplification. Note here we do not have knowledge of multiple view images $\{I_i^{C_j}\}$ belonging to the same object instance $C_j$, which makes this problem very unconstrained and challenging. We also want to learn the 3D reconstruction from a single image, i.e. $S_i = f(I_i)$, which in this paper is achieved by decomposing the mapping model into two parts $f=G \circ E$. Here $G$ is the aforementioned 3D shape generator, and $E$ represents an encoder that learns to infer shape code (or shape embedding) $z_i$ given an image $I_i$. From here on, we drop the category indicator in $I_i^C$ given no ambiguity caused.
\\

\textbf{Self-supervised Learning.} With no 3D shape data at hand, we employ the object silhouettes $\{y_i\}$ as weak supervision in a self-supervised projection-and-check manner, following recent trend \cite{yan2016perspective,kato2018neural,liu2019soft,tulsiani2017multi,yang2018learning,gwak2017weakly}. More specifically, a 3D mesh is first inferred by $S_i = G(E(I_i))$, and then re-projected to 2D with true pose parameters by a differentiable mesh renderer resulting in a projected silhouette, i.e. $x_i = R(S_i, p_i)$. Here, $x_i$ represents the re-projected silhouette of inferred shape, and $R$ represents the renderer or projector module. General differentiable mesh renderers are recently actively studied as a network  module bridging 3D mesh models and their projected 2D images \cite{loper2014opendr,kato2018neural,liu2019soft,henderson2019learning}. We adopt Kato et al.'s Neural Mesh Render (NMR) \cite{kato2018neural}, which enables back-propagation from image to mesh by making approximations to the discrete rasterization process. Then the re-projected silhouette can be checked against true silhouette, which makes the projection loss term:
\begin{equation}  \label{eqn_projLoss}
\begin{split}
\mathcal{L}^{proj}(E,G;R) & = \frac{1}{N}\sum_i^N IoU(x_i, y_i) \\
                   & = \frac{1}{N}\sum_i^N IoU(R(G(E(I_i)), p_i), y_i)
\end{split}
\end{equation}
Here, $IoU$ represents intersection-over-union metric between two binary silhouette images, \hl{and $N$ is the number of training images.} Note the neural renderer $R$ is a fixed module that does not have optimizable parameters.

Apart from the projection loss, we also add a smoothness loss term on generated meshes $S_i$, following \cite{kato2018neural,liu2019soft}:
\begin{equation}    \label{eqn_smthLoss}
\mathcal{L}^{smth}(E,G) = \frac{1}{N}\sum_i^N \sum_{e \in \mathcal{E}} (1+\cos(\theta_e))^2
\end{equation}
Where, $\mathcal{E}$ represents the set of all edges in a mesh, and $\theta_e$ is the angle between the two faces sharing an edge $e$. This smoothness term drives adjacent faces to be flat and the whole mesh model to be smooth. In this work, we use a mesh topology that consists of 642 3D vertices and 1280 triangular faces, which is initialized as a sphere.

The above self-supervised workflow is similar to previous work \cite{yan2016perspective,kato2018neural,tulsiani2017multi}. However, the difference is they assume having multiple-view observations $\{I_i^{C_j}\}$, which makes cross-view projection-and-check possible and greatly relieves 3D uncertainty. As mentioned, our setting of single-view observations $\{I_i^C\}$ makes it much harder, and we show the pose-entanglement problems and our novel treatment in the following sections.
\\

\begin{figure}[htb]
\centering
\centerline{\includegraphics[width=9cm]{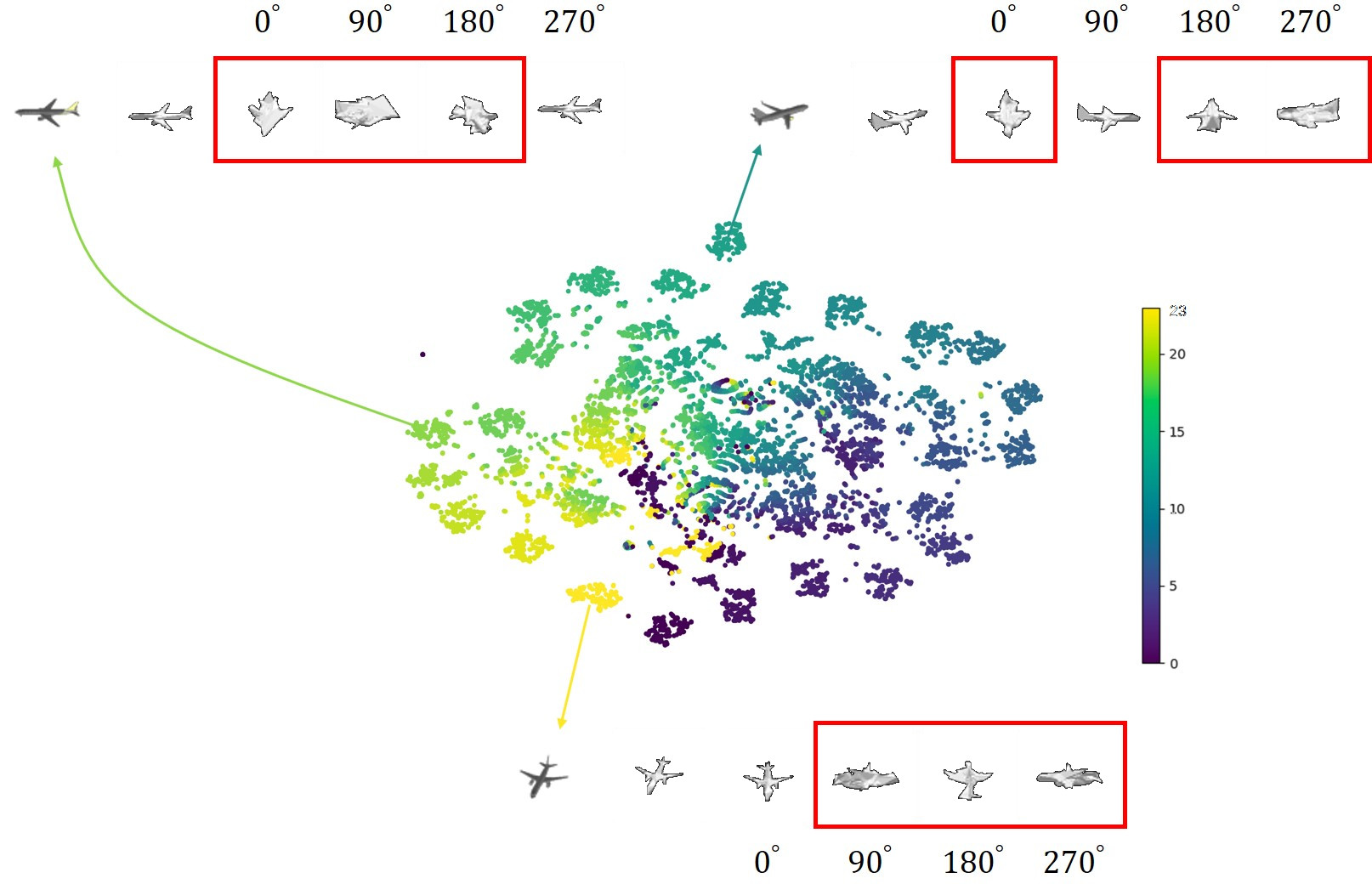}}
\caption{Distribution of shape embeddings showing the problem of pose entanglement. Embedding points are color-coded by their corresponding categorical viewpoint label of input image ($0\sim 23$). Obvious inaccuracy of three reconstructed shapes are highlighted in some viewpoints. }
\label{figEmbeding}
\end{figure}

\textbf{The Problem of Pose Entanglement.} The previous self-supervised learning strategy only works properly under multi-view supervision, while under single-view condition, the learned shape embedding has serious pose-entanglement problem. That is to say, the encoder module $E$ learns shape codes that are entangled with viewing pose of input object image. As a concrete demonstration, we show the distribution of learned shape embeddings from the vanilla auto-encoder (Vanilla-AE) model (trained with $\mathcal{L}^{proj}$ and $\mathcal{L}^{smth}$) in Fig. \ref{figEmbeding}, using t-SNE visualization method \cite{t-SNE}. The Vanilla-AE model is trained on images of airplanes taken under 24 discrete viewpoints. In Fig. \ref{figEmbeding}, shape embedding points are color-coded by their corresponding viewpoint or pose labels, and three random samples are taken showing the input image and reconstructed model rendered from input's viewpoint and four new viewpoints (i.e. $0^\circ, 90^\circ, 180^\circ, 270^\circ$).
%%%%%%%%%
%%%%%%%%%%

As can be seen, extracted shape codes $\{z_i\}$ form multiple separated clusters depending on their input images' viewpoints. The three inputs have similar shapes but are very far away in shape embedding space because of different viewpoints.
Shape reconstructions look normal under original input's viewpoint but are completely wrong under new projection directions. These observations lead to the conclusion that vanilla auto-encoder model suffers from severe pose-entanglement problem under weak single-view supervision. As a result, the direct projection-and-check strategy alone cannot guarantee to learn a single compact shape embedding space shared by all viewpoints.
\\

\textbf{Learning Pose-Disentangled Compact Shape Space.} To tackle the problem of pose-entanglement, we stress that regularization should be added on learned shape space. However, assuming no true 3D data or image correspondence information at hand,  regularizations are hard to design.
Actually, our problem is essentially very similar to unsupervised domain adaptation problems \cite{Ganin2017Domain,Tzeng2017Adversarial,Hoshen2018Unsupervised}, where the aim is to project data from different domains to the same space and align their distributions. Here, we simply consider categorical viewpoints $c_i \in \{1, 2, ..., K\}$, and treat shape codes extracted from images of different viewpoints as different domains. Since the multi-view (or multi-domain) images share the same underlying shape space, we propose to pull together distributions of shape codes from different viewpoints using adversarial domain confusion training.

A discriminator is trained on shape code and pose pairs $\{z_i, c_i\}$ to discriminate between domains (or poses). Hence pose classification loss is:
\begin{equation}    \label{eqn_clsLoss}
\mathcal{L}^{cls}(D;E) = - \frac{1}{N}\sum_i^N \log p(c_i|z_i)
\end{equation}
where, $p(c_i|z_i)$ is the softmax probability output from the discriminator network $D$.
The adversarial part tries to confuse the pose discriminator by minimizing the discrepancy between $D$'s softmax output and uniform distribution:
\begin{equation}    \label{eqn_advLoss}
\mathcal{L}^{adv}(E;D) = \frac{1}{N}\sum_i^N \sum_k^K (p(k|z_i)-\frac{1}{K})^2
\end{equation}
\hl{where $K$ is the number of categorical viewpoints.}
\hl{$\mathcal{L}^{cls}$ optimizes $D$ while fixing $E$, and $\mathcal{L}^{adv}$ optimizes $E$ while fixing $D$.
By alternatively optimizing the classification loss and adversary loss, the encoder will hopefully learn to generate shape codes that are invariant to the input image's viewpoint. 
Note that both $\mathcal{L}^{cls}$ and $\mathcal{L}^{adv}$ are applied using the same pose classifier $D$ to form a meaningful adversary, since only a well learned pose-sensitive $D$ can act as a good critic for $E$ to learn pose-invariant shape embeddings.}

Here, our adversarial training is carried out in the shape embedding space, distinguishing our method from related work using adversarial training on 3D shape space \cite{gwak2017weakly,wu2016learning,wu2018learning} or on re-projected image space \cite{gadelha20173d,yang2018learning}. We also previously tried to add adversarial loss on re-projected silhouette images from multiple new viewpoints to force regularity on cross-view projections. However, this trial fails to obtain satisfying results, likely because the image domain GAN cannot back-prop stable gradients through the complex mesh renderer, and the training eventually goes unstable and explodes. On the contrary, the proposed adversarial domain confusion training on embedding space is very stable and easy to train, and also obtains good results.

Besides the above adversarial domain confusion losses, we also propose to add a Gaussian prior regularization term on the shape embedding space. This prior comes from the intuition that instances from the same object category should be similar and hence form a compact shape space. Under Gaussian prior assumption, the regularization term is simply minimizing $L_2$ norms of shape codes:
\begin{equation}    \label{eqn_GpriorLoss}
\mathcal{L}^{prior}(E) = \frac{1}{N}\sum_i^N \|z_i \|_2
\end{equation}

Finally, we put all the loss terms together and obtain:
\begin{equation}    \label{eqn_totLoss}
\begin{split}
&\mathcal{L}^{tot}(E,G,D;R) = \mathcal{L}^{proj}(E,G;R) + \lambda_1 \mathcal{L}^{smth}(E,G) \\
&+ \lambda_2 \mathcal{L}^{cls}(D;E) + \lambda_3 \mathcal{L}^{adv}(E;D) + \lambda_4 \mathcal{L}^{prior}(E)
\end{split}
\end{equation}
where, $\lambda_1, \lambda_2, \lambda_3, \lambda_4$ are corresponding weight for each term. $\mathcal{L}^{cls}$ optimizes $D$ while fixing $E$, and $\mathcal{L}^{adv}$ optimizes $E$ while fixing $D$.
\section{Implementation Details}
The model structure is adapted from Kato et al.'s \cite{kato2018neural} with minor modifications. The encoder $E$ takes input images of size $64 \times 64$ with $4$ channels, which are three color channels and an additional channel of binary silhouette. $E$ has three convolutional layers of $5 \times 5 \times 64$, $5 \times 5 \times 128$, $5 \times 5 \times 256$, each with stride $2$ and followed by ReLU activation. After convolutional layers are two fully connected layers of dimension 1024, also with ReLU activation. The final output is another fully connected layer of dimension $512$, but we omit the ReLU that hinders the Gaussian prior on output codes distribution.

The mesh generator $G$ takes in the $512$ dimension shape code from $E$ and output a $642 \times 3 = 1926$ dimensional vector that represents 3D vertices' displacements from the initial unit sphere. $G$ is implemented simply by three fully connected layers of dimensions $1024$, $2048$ and $1926$, with the first two hidden layers followed by ReLU activation and the final output layer without ReLU. The output displacement vector is added to the sphere vertices to obtain a generated 3D mesh model.

Our discriminator $D$ is also a fully connected network with two hidden layers and a final output layer. Its input is the $512$ dimension shape code from $E$, the hidden layers are of dimensions $256, 128$ with ReLu activation, and the output layer is a $K$-way softmax layer, where $K$ is the number of discrete viewpoints in the used dataset.
We implement the proposed model using Pytorch framework, and use a Pytorch implementation\footnote{\url{https://github.com/daniilidis-group/neural\_renderer}} of NMR \cite{kato2018neural} as $R$. The NMR takes as input the generated 3D mesh model $S_i$ and a pose parameter $p_i$, which is represented by azimuth and zenith angles and a distance from camera to object center.

The weighting parameters for loss terms $\lambda_1, \lambda_2, \lambda_3, \lambda_4$ are selected as $0.001, 1, 1, 1$ respectively throughout our experiments, unless stated otherwise. For each object category, we train a model end-to-end for $20,000$ iterations with a batch size of $128$ images and a learning rate of $0.0001$. ADAM optimization algorithm is used with default parameters. We optimize either the classification loss $\mathcal{L}^{cls}$ or the adversarial loss $\mathcal{L}^{adv}$ in alternative iterations together with all the other loss terms.

\section{Experiments}
\begin{figure*}[htb]
\centering
\centerline{\includegraphics[width=13cm]{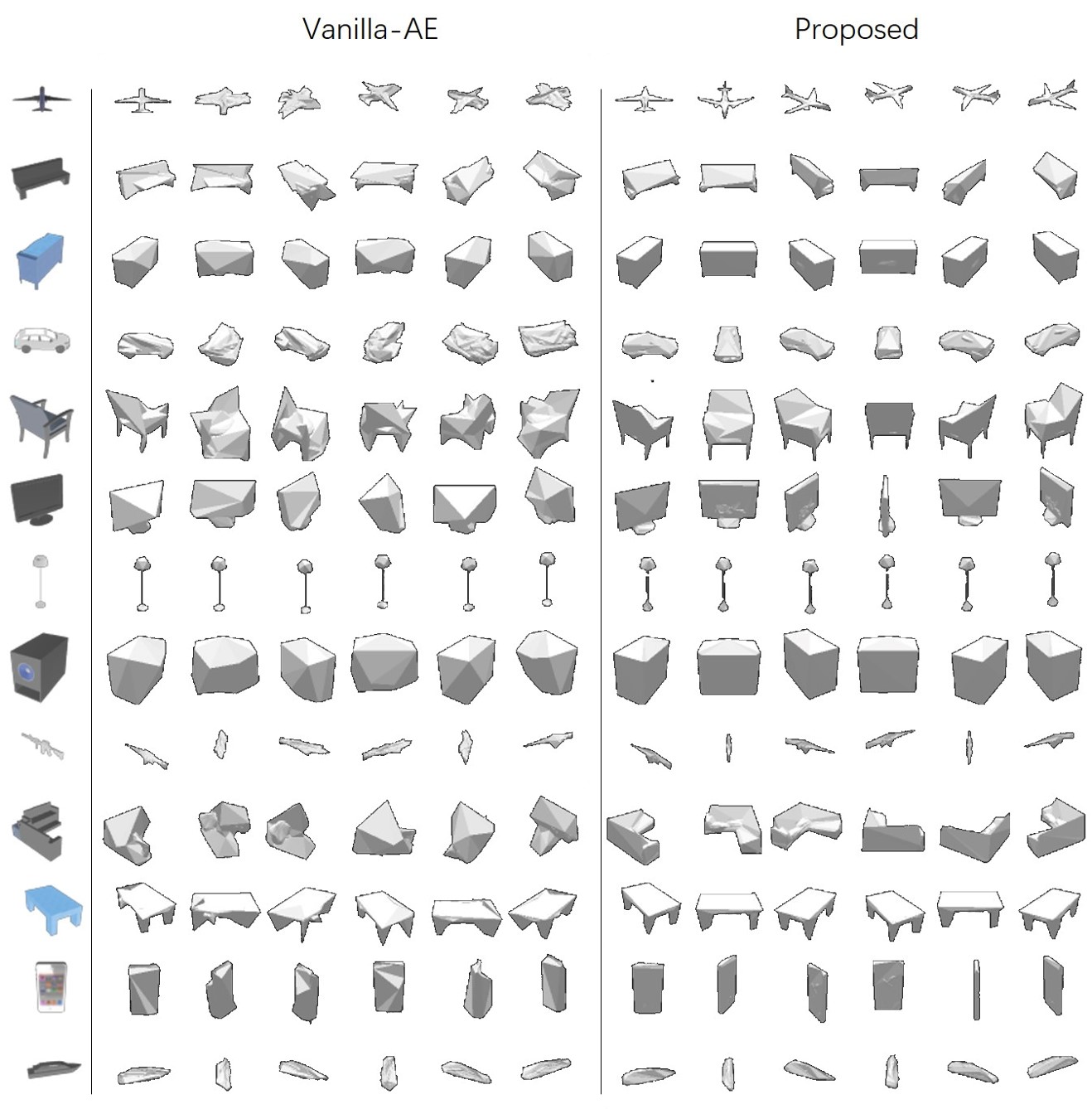}}
\caption{Comparison of reconstruction results using Vanilla-AE (left) and the proposed model (right). Input images are shown at the left-most, and reconstructed models are viewed from multiple directions.}
\label{figReconCompare}
\end{figure*}
\begin{figure*}[htb]
\centering
\centerline{\includegraphics[width=13cm]{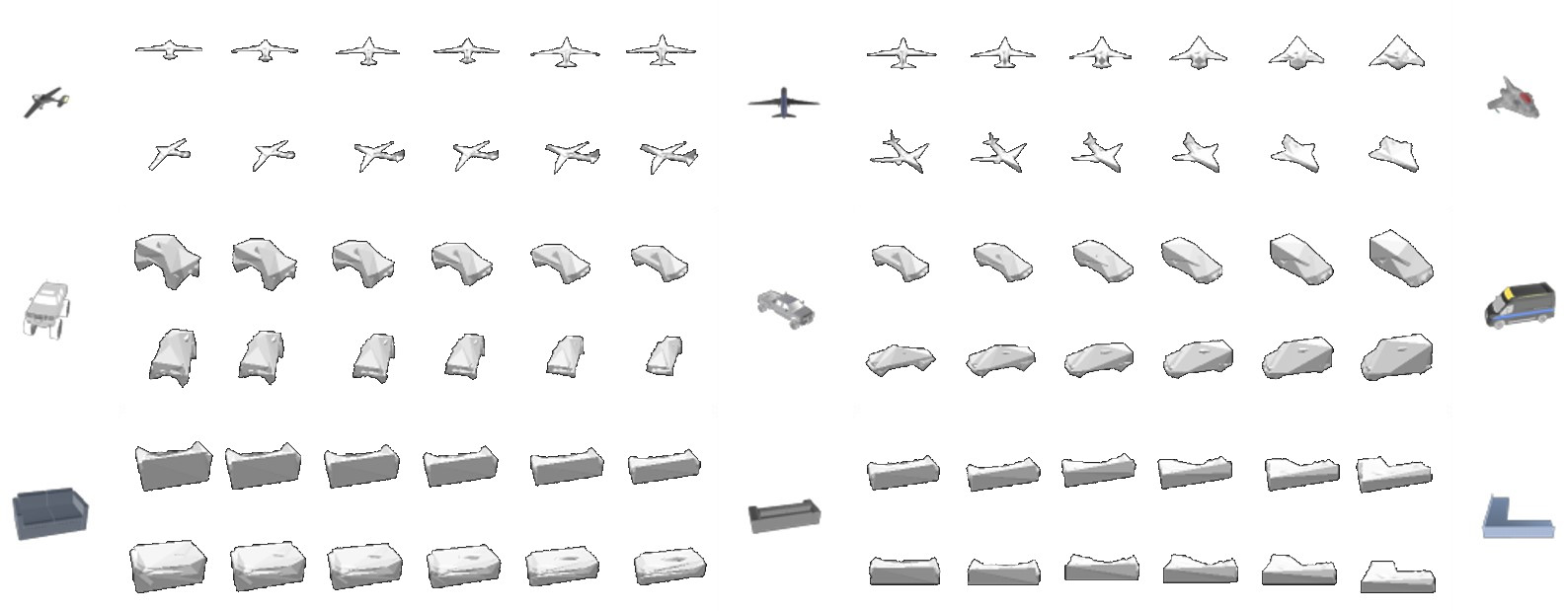}}
\caption{Shape space interpolation by the proposed model.}
\label{figInterpolation}
\end{figure*}
% %
\textbf{Datasets and Setups.} We use the synthetically rendered dataset from Kato et al. \cite{kato2018neural} and another one from Kar et al. \cite{kar2017learning} for experiments, which facilitate performance evaluation with known groundtruth 3D shapes from ShapeNet \cite{shapenet2015}. Kato's dataset \cite{kato2018neural} consists of synthetically rendered images of $13$ commonly seen object categories, all with resolution $64 \times 64$, and partitioned into non-overlapping train, validation and test sets. Each object is rendered under $K=24$ azimuthly equally spaced viewpoints around the object, on the same fixed elevation angle. Note that although each object is rendered under $24$ viewpoints, during training we discard this multi-view information and treat each image independently, making our actual training to be a single-view supervised setting. Rendered images are accompanied with corresponding groundtruth 3D voxels, which are only used for performance evaluation.

Kar's dataset \cite{kar2017learning} is also used to evaluate our method on continuously distributed viewpoints. This dataset has rendered images of the same $13$ object categories, with each object instance rendered under $20$ randomly sampled viewpoints in the range of $360^{\circ}$ azimuth and $-20^{\circ} \sim 40^{\circ}$ elevation angles. To adapt our method on this dataset, the continuous viewpoints are discretized into $72$ bins in granularity of $(15^{\circ}, 20^{\circ})$ in azimuth and elevation. As a result, the discriminator in our model has $K = 72$ softmax outputs for this dataset. Since the VPL method \cite{kato2019learning} also conducts experiments on this dataset, we use the same setting as in \cite{kato2019learning} for comparison. Following \cite{kato2019learning}, only a single-view image is sampled from the $20$ images of each object instance for training.
\\

\textbf{Qualitative Results of Shape Reconstruction.} We first qualitatively evaluate 3D shape reconstruction of the proposed model, using Kato's dataset \cite{kato2018neural}. We compare with the vanilla Auto-encoder (Vanilla-AE) model proposed in \cite{kato2018neural}, which is the base model that our method is developed upon. Different from \cite{kato2018neural}, where Vanilla-AE model is trained using multi-view images, we retrain the model in our single-view setting.

In Fig. \ref{figReconCompare}, we show a qualitative comparison of reconstruction results by Vanilla-AE and proposed model. One testing sample image from each of $13$ categories is input to the two models, and output 3D meshes are rendered in the original viewpoints and $5$ new viewpoints to show the reconstruction quality. We can see clear pose entanglement problem for the Vanilla-AE reconstructions. Most of its reconstructed shapes have faithful silhouettes under the original viewpoints of input images, but are far from normal shapes once seen from other new viewpoints. On the contrary, reconstructions from the proposed model show faithful and consistent shapes under all viewpoints. This comparison intuitively illustrates the effectiveness of proposed domain confusion training to learn a pose-disentangled and compact shape space. Note that the proposed method has trouble reconstructing the concave parts in objects like bench, chair and sofa, as can be seen from Fig. \ref{figReconCompare}. This is an inherent deficiency of silhouette based reconstruction methods, because concave parts cannot be expressed by silhouettes under all viewpoints, leading to the best guess of object's visual hull.
\\

\begin{table*} [htb] 
\centering
\caption{Comparison of shape reconstruction accuracy using voxel IoU on Kato's dataset \cite{kato2018neural}.}  \label{tabReconCampare}
\renewcommand{\arraystretch}{1.0}
\scalebox{1.0}{%
\begin{tabular}{ c c c c c c c c c}
  \toprule
                    & Airplane   & Bench   & Dresser    & Car   &Chair      &Display    &Lamp  \\
  \hline
  Vanilla-AE \cite{kato2018neural}			 &0.318    	 &0.304    &0.377  	  &0.534  	&0.233  	 &0.267  		&0.378    \\
  \hline
 VPL \cite{kato2019learning}
							 &0.533   	 &0.367  	&\textbf{0.666}   	  &0.635  	&0.386  	 &0.500 		&0.330   \\
  \hline
  Proposed      	 &\textbf{0.565}    	 &\textbf{0.410}  	&0.661  	  &0.670  	&\textbf{0.415}  	 &\textbf{0.532}  		&0.378    \\
  \hline
  \hline
                    & Loudspeaker   & Rifle   & Sofa    & Table   &Telephone      &Vessel  &  &\textbf{Mean}  \\
  \hline
  Vanilla-AE \cite{kato2018neural}   		 &0.287       	 &0.333   	&0.390  	&0.375    &0.376  		  &0.475  	&  	&0.357    \\
 \hline
	VPL \cite{kato2019learning}
							 &\textbf{0.576}      	 &0.356  	&0.558  	&0.416	&0.670     	  &0.453  & 			&0.496  \\
  \hline
  Proposed      	 &0.434      	 &\textbf{0.533} 	&\textbf{0.592}	&\textbf{0.417}    &0.684 		  &\textbf{0.528} 	&  		&\textbf{0.524}     \\
  \bottomrule
\end{tabular}
}
\end{table*}

\textbf{Shape Space Interpolation.}
In Fig. \ref{figInterpolation}, we demonstrate the effects of shape space interpolation. Three sample images are projected to shape embedding space by encoder $E$ and intermediate shape codes are linearly interpolated between them and reconstructed to 3D models by generator $G$. Reconstructed shapes are viewed under two viewpoints. As can be seen from the demonstration, the morphing process of 3D shapes is smooth and realistic, implying a compact shape space has been learned. The proposed model also successfully learned diverse multi-mode characteristic of shape space, like regular and composite sofas, business airliners and jet fighters and different kinds of cars.

\textbf{Quantitative Comparison on Kato's Synthetic Dataset \cite{kato2018neural}}
We evaluate quantitative reconstruction accuracy using the intersection-over-union (IoU) metric between voxelized true and reconstructed 3D mesh models on the test set. \hl{Here, we use voxelization of $32 \times 32 \times 32$.} The mean IoUs for each object category and their overall mean are listed in Table \ref{tabReconCampare} for Kato's dataset \cite{kato2018neural}. Compared methods are Vanilla-AE model \cite{kato2018neural} trained with single-view setting, the VPL model in \cite{kato2019learning} and the full proposed method. The VPL model also uses adversarial training, but is conducted in the re-projected image domain to make re-projected images from input view and unobserved view indistinguishable. We implemented the most related view prior learning part in \cite{kato2019learning} using the same backbone network as ours, leaving their irrelevant texture prediction and internal pressure components.
All models are trained for $20,000$ iterations and the final models are tested on test sets.

As can be seen from Table \ref{tabReconCampare}, our overall improvement is nearly $17$ points compared to Vanilla-AE model, clearly proving the effectiveness of our method under the scenario of single-view training.
Compared to the VPL method \cite{kato2019learning}, our accuracy is also better. Besides, the proposed method is much more efficient than VPL, where our model only takes 2 hours for the whole 20K training iterations on a NIVIDIA Titan-Xp GPU, while VPL requires 4$\sim$5 hours under the same conditions. Comparing the two methods, our work conduct domain-confusion in shape embedding space, which is intuitively more direct and simpler compared to VPL’s domain-confusion in re-projected image space. This is because their domain-confusion effect has to be back-propped from re-projected image to reconstructed shape and then to shape embedding space, thus has two more modules to back-propagate compared to our method that directly treat the essential shape space. Thus our method can achieve on-par or better reconstruction accuracy and is much more efficient.

\begin{table} [htb] 
\centering
\caption{\hl{Comparison with the PrGAN method \mbox{\cite{gadelha20173d}} using voxel IoU on Kato's dataset \mbox{\cite{kato2018neural}}.}}  \label{tabComparePrGAN}
\renewcommand{\arraystretch}{1.0}
\scalebox{1.0}{%
\begin{tabular}{ c c c c c}
  \toprule
                    & Airplane   & Car   &Chair      &Mean  \\
  \hline
  PrGAN \cite{gadelha20173d}	&0.151    	 &0.228    &0.114  	  &0.164    \\
  \hline
  Proposed      	 &\textbf{0.565}   &\textbf{0.670}  &\textbf{0.415}  	 &\textbf{0.550}   \\
  \bottomrule
\end{tabular}
}
\end{table}
\hl{Apart from the VPL method, we also compare with PrGAN proposed in \mbox{\cite{gadelha20173d}}, which is another method that can be trained on single-view image dataset to learn 3D object model. PrGAN is a GAN model that learns to generate 3D voxels from noise samples (latent codes), and a discriminator is trained in the re-projected silhouette image domain to guide the 3D generator. PrGAN does not have an encoder model that maps input image to latent shape code. To compare with it, we follow the original paper \mbox{\cite{gadelha20173d}} and train an encoder post-hoc using pairs of noise sample and its generated silhouette image to learn the inverse mapping. We adapt PrGAN's official TensorFlow code to conduct experiment on Kato's synthetic dataset. The results can be seen from Table \mbox{\ref{tabComparePrGAN}}, where only three representative object classes are tested, because PrGAN training is very slow, taking 2 days for a single class. As can be seen, the proposed accuracy is a lot better than PrGAN. Apart from the effectiveness of our method, a part of the reason is that PrGAN is mainly proposed for unconditioned 3D generation but not 3D reconstruction, so its inference capability from image to 3D space is not directly learned and results are not very good.}

\textbf{Quantitative Comparison on Kar's Synthetic Dataset \cite{kar2017learning}}. We also compare with the VPL method on Kar's dataset \cite{kar2017learning} in Table \ref{tabReconCampare2} to see the performance on continuously distributed viewpoint images. Compared methods are the baseline vanilla-AE model \cite{kato2018neural} and the VPL model \cite{kato2019learning} with their performances reported in \cite{kato2019learning} on this dataset. For the VPL model, we select their reported performance with the configuration of both view-prior-learning and internal-pressure on, but without texture prediction \hl{(the 3rd row of Table 2 in \mbox{\cite{kato2019learning}})}. Since our model is trained for each object category, we also compare the VPL model that is trained with class conditioning for more fair comparison. Also, as with \cite{kato2019learning}, we select our model with the best validation accuracy for the final test. As can be seen from Table \ref{tabReconCampare2}, our method performs much better than baseline Vanilla-AE, illustrating the effectiveness of proposed domain confusion training for pose-disentanglement. Although our accuracy is marginally shy compared with VPL, the numbers are on-par. Given that VPL uses a more advanced neural mesh renderer and stronger backbone models, and the fact that our model is much more efficient than VPL, \hl{the proposed method can be considered as very competitive.}
\begin{table*} [htb] 
\centering
\caption{Comparison of shape reconstruction accuracy using voxel IoU on Kar's dataset \cite{kar2017learning}.}  \label{tabReconCampare2}
\renewcommand{\arraystretch}{1.0}
%\scalebox{1.0}{%
\begin{tabular}{ c c c c c c c c c}
  \toprule
                    & Airplane   & Bench   & Dresser    & Car   &Chair      &Display    &Lamp  \\
  \hline
  Vanilla-AE \cite{kato2018neural}    &0.479 &0.266 &0.466 &0.550 &0.367 &0.265 &\textbf{0.454}  \\
  \hline
  VPL \cite{kato2019learning}    &0.513 &\textbf{0.376} &\textbf{0.591} &0.701 &\textbf{0.444} &\textbf{0.425} &0.422  \\
  \hline
  proposed     &\textbf{0.528}   &0.360 &0.569  &\textbf{0.742} &0.433 &0.423  &0.378    \\
  \hline
    \hline
                    & Loudspeaker   & Rifle   & Sofa    & Table   &Telephone      &Vessel &   &\textbf{Mean}  \\
  \hline
  Vanilla-AE \cite{kato2018neural}    &0.524 &0.382 &0.367 &0.342 &0.337 &0.439 & &0.403  \\
  \hline
  VPL \cite{kato2019learning}    &\textbf{0.596} &0.479 &0.500 &\textbf{0.436} &0.595 &0.485 & &\textbf{0.505}    \\
  \hline
  proposed     &0.528     &\textbf{0.542} &\textbf{0.537} &0.384 &\textbf{0.624}    &\textbf{0.505} & &0.504  \\
  \bottomrule
\end{tabular}
\end{table*}

\hl{\textbf{Ablation Study of Losses.} We also do ablation study on our method with one loss term turned off at one time. From Table \mbox{\ref{tabAblation}} we can see that, without the adversarial domain confusion loss $\mathcal{L}^{adv}$ the accuracy drops significantly by $10$ points. The Gaussian prior loss $\mathcal{L}^{prior}$ on embedding space also has positive effects, although not as much as adversarial loss. It seems surprising that without mesh smoothness loss $\mathcal{L}^{smth}$, the accuracy can be further improved. However, we find that these reconstructed meshes tend to have much more self-intersections and messy interior, making the reconstruction results less regularized. As a result, we stick with using smoothness loss to obtain more regular meshes.}
\begin{table} [htb] \small
\centering
\caption{\hl{Ablation study of used loss terms. Listed are mean voxel IoUs on Kato's dataset \mbox{\cite{kato2018neural}}.}}  \label{tabAblation}
\renewcommand{\arraystretch}{1.0}
\scalebox{1.0}{%
\begin{tabular}{ c c c c }
  \toprule
    Proposed   &w/o $\mathcal{L}^{prior}$   &w/o $\mathcal{L}^{adv}$    &w/o $\mathcal{L}^{smth}$   \\
  \hline
    0.524     &0.507    &0.420   &0.541    \\
  \bottomrule
\end{tabular}
}
\end{table}
%
% \begin{table*} [htb] 
% \centering
% \caption{Ablation study of used loss terms. Listed are voxel IoUs on Kato's dataset \cite{kato2018neural}.}  \label{tabAblation}
% \renewcommand{\arraystretch}{1.0}
% \scalebox{1.0}{%
% \begin{tabular}{ c c c c c c c c c}
%   \toprule
%                     & Airplane   & Bench   & Dresser    & Car   &Chair      &Display    &Lamp  \\
%   \hline
%   Proposed      	 &\textbf{0.565}    	 &\textbf{0.410}  	&0.661  	  &0.670  	&\textbf{0.415}  	 &0.532  		&0.378    \\
%   \hline
%   w/o $\mathcal{L}^{prior}$     	 &0.552    	 &0.350   	&0.581   	  &\textbf{0.680}   &0.313  	 &0.508   	&0.385    \\
%   \hline
%   w/o $\mathcal{L}^{adv}$     	 &0.403  &0.371 &0.553  &0.572  &0.329  &0.369  &0.376     \\
%   \hline
%   w/o $\mathcal{L}^{smth}$     	 &0.559  &0.386 &\textbf{0.666}  &0.654  &0.401  &\textbf{0.543}  &\textbf{0.410}    \\
%   \hline
%   \hline
%                     & Loudspeaker   & Rifle   & Sofa    & Table   &Telephone      &Vessel  &  &\textbf{Mean}  \\
%   \hline
%   Proposed      	 &0.434      	 &0.533 	&0.592	&\textbf{0.417}    &0.684 		  &\textbf{0.528} 	&  		&0.524     \\
%  \hline
%   w/o $\mathcal{L}^{prior}$  	 &0.500      	 &0.519 	&0.583  	&0.413    &0.687    		  &0.515 	& 		&0.507            \\
%   \hline
%   w/o $\mathcal{L}^{adv}$    	 &0.335  &0.352  &0.500  &0.385  &0.4350  &0.475  & &0.420     \\
%   \hline
%   w/o $\mathcal{L}^{smth}$     	 &\textbf{0.618}  &\textbf{0.588}  &\textbf{0.628}  &0.385  &\textbf{0.723}   &0.476  & &\textbf{0.541}    \\
%   \bottomrule
% \end{tabular}
% }
% \end{table*}

\hl{\textbf{Effect of the Number of Viewpoint Bins.} For datasets with continuous viewpoints, e.g. Kar's dataset \mbox{\cite{kar2017learning}}, the effect of different choices of viewpoint bins $K$ is investigated in this section. In the above quantitative experiments (Table \mbox{\ref{tabReconCampare2}}), we selected azimuth and elevation bin widths to be $15^\circ$ and $20^\circ$ respectively. In the following, we test different bin widths for Kar's dataset to see its effect on final reconstruction result. As can be seen from Table \mbox{\ref{tabViewBins}}, our method is not very sensitive to the choice of viewpoint bins, reflecting the robustness of the domain confusion term to hyper-parameter tuning. More specifically, our choice of $(15^\circ,20^\circ)$ in the above experiment in Table \mbox{\ref{tabReconCampare2}} corresponds to the best performance among tested viewpoint binning choices.}
\begin{table} [htb] \small
\centering
\caption{\hl{Effect of the number of viewpoint bins $K$ on mean reconstruction IoU of Kar's dataset \mbox{\cite{kar2017learning}}. For each column, the bin widths of azimuth and elevation are respectively: $(60^\circ,60^\circ)$, $(30^\circ,30^\circ)$, $(20^\circ,20^\circ)$, $(20^\circ,15^\circ)$, $(15^\circ,20^\circ)$, $(15^\circ,15^\circ)$, $(10^\circ,10^\circ)$.}}  \label{tabViewBins}
\renewcommand{\arraystretch}{1.0}
\scalebox{0.8}{%
\begin{tabular}{ c c c c c c c }
  \toprule
                 $K=6$   & $K=24$   & $K=54$    & $K=72$   & $K=72$     & $K=96$    & $K=216$  \\
  \hline
  0.497     &0.499    &0.501   &0.500    &0.504     &0.498     &0.490    \\
  \bottomrule
\end{tabular}
}
\end{table}

\begin{figure}[htb]
\centering
\centerline{\includegraphics[width=8.5cm]{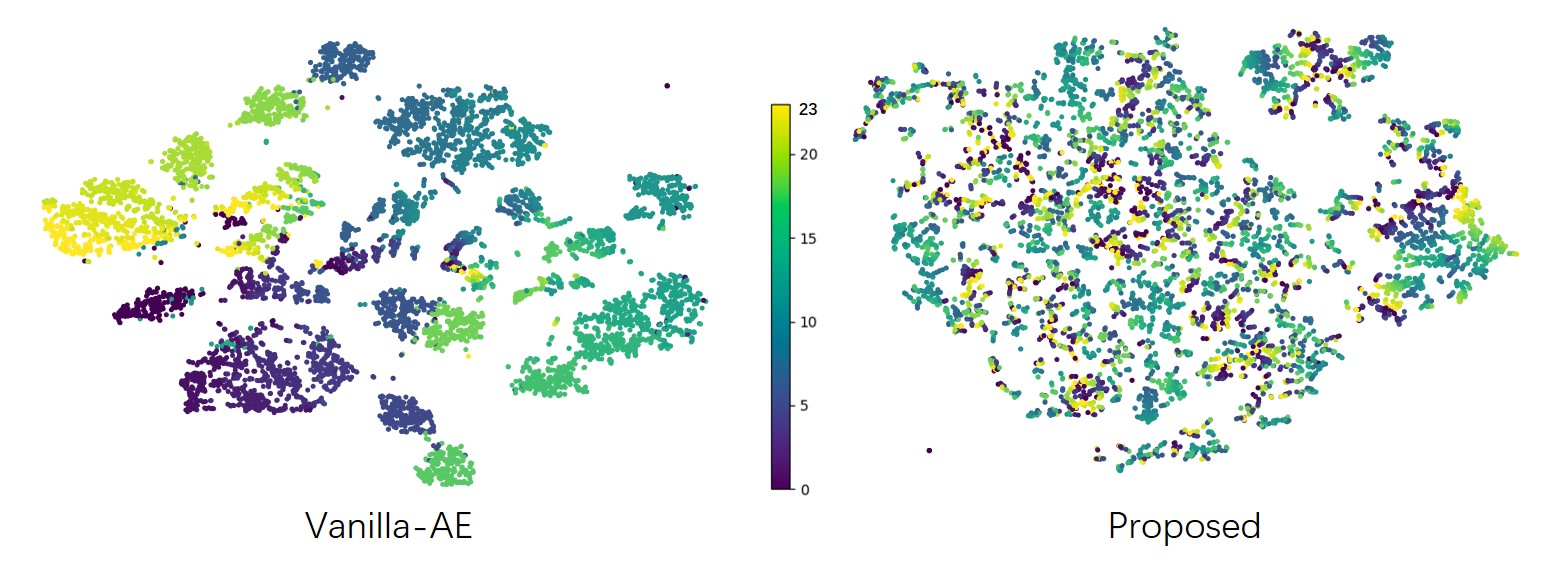}}
\caption{Comparison of shape embedding distributions for the \textit{sofa} category. Points are color-coded by pose category.}
\label{figEmbedingComp}
\end{figure}

\textbf{Effect of Domain Confusion Training.} In the following, we demonstrate the influence of proposed method on learned shape embedding space both qualitatively and quantitatively. As an example, the comparison of shape embedding distributions using t-SNE for \textit{sofa} test images in Kato's dataset are shown in Fig. \ref{figEmbedingComp}. Without domain confusion training, the Vanilla-AE model's shape codes are clearly entangled with input poses, showing separate clusters of different color. On the contrary, our model successfully produces pose-invariant shape codes that are more mixed together. This example validates the effectiveness of introducing domain confusion losses on learning pose-disentangled consistent  shape spaces.
\begin{table*} [htb] \small
\centering
\caption{Comparison of Maximum Mean Discrepancy (MMD) distances between shape codes extracted under different viewpoints. Smaller values indicate better pose-invariance of learned shape space.}  \label{tabMMD}
\renewcommand{\arraystretch}{1.0}
%\scalebox{1.0}{%
\begin{tabular}{ c c c c c c c c c}
  \toprule
                    & Airplane   & Bench   & Dresser    & Car   &Chair      &Display    &Lamp  \\
  \hline
  Vanilla-AE    &1.825     &2.389    &0.737   &4.040    &1.240     &1.488     &0.018    \\
  \hline
  proposed     &0.272     &0.382     &0.051     &0.990     &0.106       &0.152     &0.014    \\
  \hline
    \hline
                    & Loudspeaker   & Rifle   & Sofa    & Table   &Telephone      &Vessel &   &\textbf{Mean}  \\
  \hline
  Vanilla-AE    &0.647    &2.385    &2.309    &0.521    &2.306     &1.848  &   &1.673    \\
  \hline
  proposed     &0.079     &0.249     &0.173     &0.033     &0.298       &0.639  &   &0.264    \\
  \bottomrule
\end{tabular}
\end{table*}

As a quantitative measure of the effect of domain confusion training, we also use Maximum Mean Discrepancy (MMD) \cite{MMD} and calculate distances between shape codes extracted under different viewpoints of Kato's dataset. MMD is an effective metric for \hl{measuring the distance between two distributions $X, Y$ each of which has $m$ and $n$ samples:}
\begin{small}
\begin{equation}  \label{eqn_MMD}
\begin{split}
    & MMD(X, Y) = \\
    & \left[ \frac{1}{m^2} \sum_{i,j=1}^{m} k(x_i, x_j) - \frac{2}{mn} \sum_{i,j=1}^{m,n} k(x_i, y_j) + \frac{1}{n^2} \sum_{i,j=1}^{n} k(y_i, y_j)  \right]^{\frac{1}{2}}
\end{split}
\end{equation}
\end{small}
\hl{, and here we use Gaussian RBF kernel for $k(\cdot,\cdot)$.
More specifically, for each object class we sample $512$ test images ($m=n$) for each of $24$ viewpoints and obtain their shape codes using encoder $E$, MMDs are calculated between code samples from all possible pairs of different viewpoints. There are totally $C_{24}^2 = 276$ combinational conditions (i.e. 276 pairs of $(X,Y)$ combinations for Eqn. \mbox{\ref{eqn_MMD}}) and the mean of these MMDs is taken as the final overall distance metric.}
The mean MMD metrics for all object categories are listed in Table \ref{tabMMD}, where smaller values indicate closer distances between different viewpoints and hence better pose-invariance quality. It shows that MMDs of our proposed model are much smaller than those of Vanilla-AE model for all $13$ object categories. On average, our distance is only $15\%$ of Vanilla-AE model's. This quantitative comparison concretely proves the effectiveness of proposed domain confusion training to pull together shape embeddings from different viewpoints.

\hl{\textbf{Experiment on real-world dataset.}} Apart from the above experiments on two synthetic datasets, we also conduct experiments on a real-world dataset, i.e. Pascal 3D+ \cite{xiang2014beyond, tulsiani2017multi}, which includes annotations for approximate 3D models, viewpoints and silhouettes. We used a compilation of this dataset from Kato et al. \cite{kato2019learning} and followed their setting, where we run our method five times with different random initialization and report the mean IoU accuracy. Since the viewpoints in this dataset are also continuous, similar to Kar's synthetic dataset \cite{kar2017learning}, we again discretize viewpoints to categorical bins. For Pascal 3D+ dataset, we use bin width of $45^{\circ}$ for discretizing viewpoints and set $\lambda_1, \lambda_2, \lambda_3, \lambda_4$ to $0.001, 1, 0.01, 0.01$ respectively. To adapt to the $224 \times 224 \times 3$ sized input images in Pascal 3D+, we choose the ResNet-18 model \cite{he2016deep} as our encoder. Following previous work \cite{kanazawa2018learning, kato2019learning}, we also constrain our mesh generator to output symmetric 3D models.

\begin{figure}[htb]
\centering
\centerline{\includegraphics[width=8.5cm]{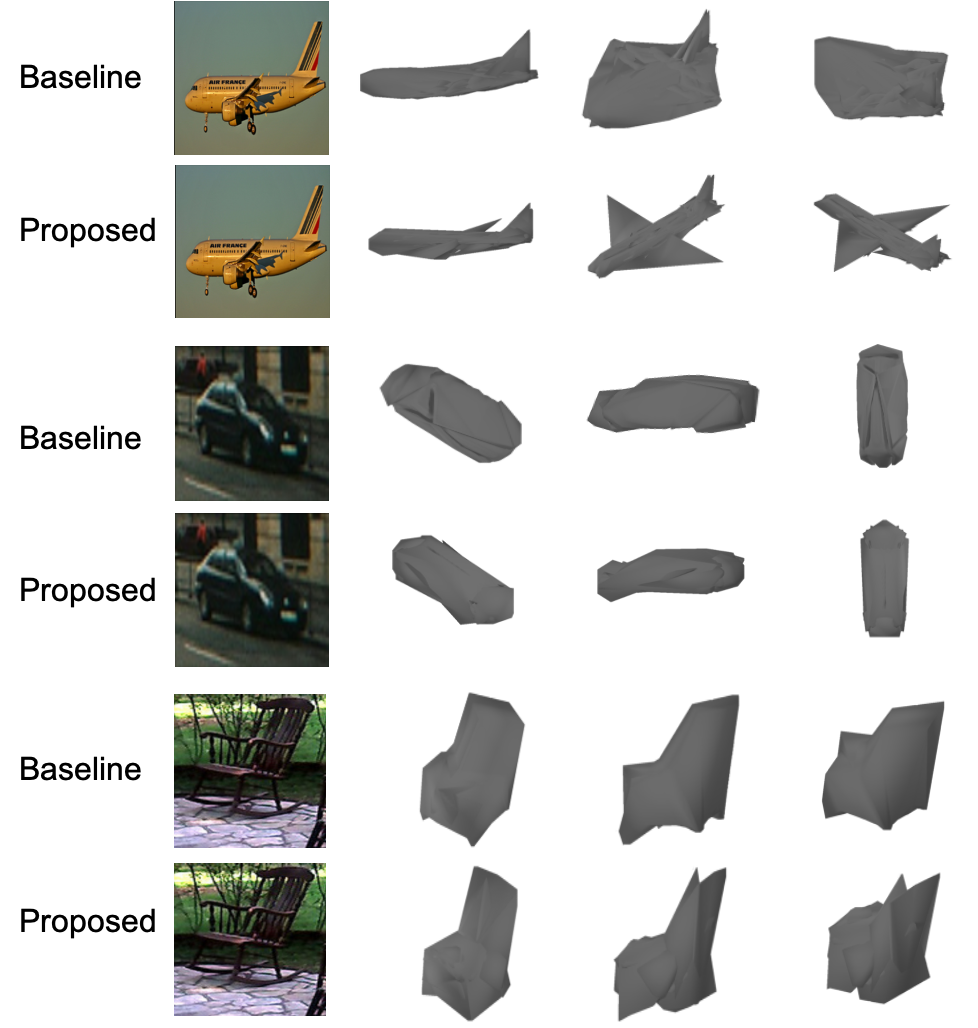}}
\caption{\hl{Comparison of reconstruction results on Pascal 3D+ images using the baseline method (i.e. Vanilla-AE) and the proposed method. The left-most images are input images and the rest columns are reconstructed models viewed from different directions.}}
\label{figPascalRecon}
\end{figure}

The quantitative and qualitative reconstruction results on Pascal 3D+ are shown in Table \ref{tabComparePascal3D} and Fig. \ref{figPascalRecon} respectively. As can be seen from these results, the proposed method surpassed the baseline (i.e. the Vanilla-AE method \cite{kato2018neural}) both quantitatively and qualitatively, proving the efficacy of proposed pose-disentanglement training on real-world images. More specifically, the proposed method improves IoU score by 2 points over the baseline. From Fig. \ref{figPascalRecon}, our reconstructed airplane has more clear wings and the chair has thinner back. From Table \ref{tabComparePascal3D}, it can be seen our method also surpassed two previous methods CSDM \cite{kar2015category} and CMR \cite{kanazawa2018learning} which are also in single-view training setting. However, our accuracy is a little worse than the VPL \cite{kato2019learning} method by 1 point in average. Here we list the VPL accuracy as reported in its original paper \cite{kato2019learning} without texture prediction. Analyzing respective results of three object categories, our method has better accuracy for mostly convex objects, e.g. cars, but worse results for concave objects, e.g. chair and airplane, compared to VPL. We note that the difference in accuracy is not very large, and the ``groundtruth" 3D models are actually approximate sudo-groundtruth in this real-world dataset. Also considering that our method is more efficient and simpler in design, we say that the proposed method is competitive with state-of-the-art and very promising for real-world application.
\begin{table} [htb]
\centering
\caption{\hl{Comparison of shape reconstruction accuracy using voxel IoU on Pascal 3D+ real-world dataset \mbox{\cite{xiang2014beyond}}.}}  \label{tabComparePascal3D}
\renewcommand{\arraystretch}{1.0}
\begin{tabular}{ c c c c c }
  \toprule
                 & Airplane   & Car     & Chair   & Mean \\
  CSDM \cite{kar2015category}          &0.398       &0.600    &0.291    &0.429    \\
  CMR \cite{kanazawa2018learning}          &0.46       &0.64    &n/a    &n/a    \\
  VPL \cite{kato2019learning}          &\textbf{0.472}       &0.689    &\textbf{0.303}    &\textbf{0.488}    \\
  \hline
  Baseline \cite{kato2018neural}     & 0.423    &0.672  &0.265  &0.453 \\
  Proposed      &0.458   &\textbf{0.695}  &0.277  &0.477 \\
  \bottomrule
\end{tabular}
\end{table}

\section{Conclusions}
In this paper, we investigate the challenging problem of learning high quality 3D generative models from only single-view images. Compared to previous major settings of using multiple view images as supervision, single-view setting is a more loose and practical assumption, but has serious challenge of under-constraint. This leads to the problem of pose entanglement, where shapes reconstructed under different viewpoints are greatly divergent. To address this problem and learn a pose-invariant reconstruction model, we cast it to a domain adaptation problem by treating shape embeddings from different viewpoints as different domains, and propose an adversarial domain confusion training method. Comprehensive experiments are conducted showing the effectiveness of proposed model.

This work is among the first attempts of employing deep learning methods to achieve faithful 3D mesh reconstruction using only single-view image supervision. Although the results are rather promising, we note there are still some problems remaining to be tackled in the future. Firstly, improving reconstruction performances to approach those of multi-view supervision requires more study. \hl{Secondly, this work assumes known poses, which cannot always be easily acquired or estimated beforehand. Developing methods that can simultaneously estimate both pose and 3D shape using only single-view images is a much more challenging problem that needs future efforts.}
% Our initial attempts fail to obtain meaningful results for this most hard setting, which we think needs more efforts.
\hl{Last but not least, considering reconstruction uncertainty is another important issue, as single-view reconstruction usually corresponds to multiple possible 3D solutions. In this work, we only obtain a single deterministic solution by the Auto-Encoder. In future work, probabilistic models such as Variational Auto-Encoders (VAE) \mbox{\cite{kingma2013auto}} can be studied to better model the distribution of 3D reconstructions.}

%% Loading bibliography style file
%\bibliographystyle{model1-num-names}
\bibliographystyle{cas-model2-names}

% Loading bibliography database
\bibliography{egbib}

%\vskip3pt

\bio{}
Author biography without author photo.
Author biography. Author biography. Author biography.
Author biography. Author biography. Author biography.
Author biography. Author biography. Author biography.
Author biography. Author biography. Author biography.
Author biography. Author biography. Author biography.
Author biography. Author biography. Author biography.
Author biography. Author biography. Author biography.
Author biography. Author biography. Author biography.
Author biography. Author biography. Author biography.
\endbio

\bio{figs/pic1}
Author biography with author photo.
Author biography. Author biography. Author biography.
Author biography. Author biography. Author biography.
Author biography. Author biography. Author biography.
Author biography. Author biography. Author biography.
Author biography. Author biography. Author biography.
Author biography. Author biography. Author biography.
Author biography. Author biography. Author biography.
Author biography. Author biography. Author biography.
Author biography. Author biography. Author biography.
\endbio

\bio{figs/pic1}
Author biography with author photo.
Author biography. Author biography. Author biography.
Author biography. Author biography. Author biography.
Author biography. Author biography. Author biography.
Author biography. Author biography. Author biography.
\endbio

\end{document}